\documentclass[english]{article}
\usepackage[T1]{fontenc}
\usepackage[latin9]{inputenc}
\usepackage{array}
\usepackage{booktabs}
\usepackage{textcomp}
\usepackage{url}
\usepackage{bm}
\usepackage{multirow}
\usepackage{amsmath}
\usepackage{amssymb}
\usepackage{graphicx}
\usepackage{esint}
\usepackage[numbers]{natbib}
\makeatletter

\providecommand{\tabularnewline}{\\}

\usepackage{fullpage}
\usepackage{bm}
\usepackage{booktabs}

\DeclareMathOperator*{\argmax}{arg\,max}

\makeatother

\usepackage{babel}
\begin{document}

\title{Density Ratio Hidden Markov Models}

\author{John A. Quinn\\
Department of Computer Science \\
Makerere University\\
PO Box 7062, Kampala, Uganda\\
\url{jquinn@cit.ac.ug} \and Masashi Sugiyama\\
Department of Computer Science\\
Tokyo Institute of Technology\\
Tokyo 152-8552, Japan\\
\url{sugi@cs.titech.ac.jp}}

\date{~}
\maketitle
\begin{abstract}
Hidden Markov models and their variants are the predominant sequential
classification method in such domains as speech recognition, bioinformatics
and natural language processing. Being generative rather than discriminative
models, however, their classification performance is a drawback. In
this paper we apply ideas from the field of density ratio estimation
to bypass the difficult step of learning likelihood functions in HMMs.
By reformulating inference and model fitting in terms of density ratios
and applying a fast kernel-based estimation method, we show that it
is possible to obtain a striking increase in discriminative performance
while retaining the probabilistic qualities of the HMM. We demonstrate
experimentally that this formulation makes more efficient use of training
data than alternative approaches.
\end{abstract}

\section{Introduction}

Inference of a sequence of estimated classes from a sequence of noisy
observations is fundamental in many applications. The hidden Markov
model (HMM) and its variants are the usual methods employed to do
this, and have been used with conspicuous success in such domains
as speech recognition, bioinformatics and natural language processing.
As well as being computationally efficient, they are a popular choice
due to their intuitive probabilistic interpretation. However, they
have drawbacks in terms of classification accuracy, being primarily
generative rather than discriminative models.

One established approach to improve classification performance in
HMMs models has been to adapt the model to discriminative forms which
apply information theoretic principles in training \cite{mccallum2000maximum,woodland2002large}.
This improves classification performance, though the requirement in
general to limit such models to parametric forms means they still
do not have the discriminative power of kernel-based and max-margin
methods. Another idea therefore is to create models combining the
structure of the HMM and the classification approach of Support Vector
Machines \cite{altun03hiddenSVM,koller2004maxmargin,tsochantaridis2006large}.
This considerably improves classification performance, though at the
expense of losing the intuitive probabilistic interpretation of the
HMM.

In this paper, we propose a different idea which combines the advantages
of both the above approaches, using concepts from the field of direct
density ratio estimation \cite{book:Sugiyama+etal:2012,IEEE-IT:Nguyen+etal:2010,ICML:Bickel+etal:2007}.
Our key observation is that rather than trying to quantify a set of
\emph{likelihoods} (i.e. how likely any possible observation is given
some class, relative to the likelihood of making other observations
given that same class), it is in fact only necessary to know about
\emph{likelihood ratios} (i.e. how likely any possible observation
is given some class, relative to the likelihood of making the same
observation given a different class). Because the forward-backward
inference algorithm computes such ratios anyway, we therefore follow
the principle that we should avoid solving a more general learning
problem than is strictly necessary \cite{vapnik98learningtheory}.

We therefore reformulate the forward-backward algorithm for HMM inference
in terms of density ratios so that the intermediate step of likelihood
function estimation can be dispensed with. We demonstrate how efficient
and highly discriminative nonparametric inference can be carried out
in this framework using a kernel-based density ratio estimation procedure
\cite{JMLR:Kanamori+etal:2009}. Because density ratios are a natural
parameterization of the forward-backward algorithm, the resulting
inference procedure is also more numerically stable than the conventional
version. We demonstrate these ideas using synthetic data and on a
physiological monitoring problem. In the case of physiological monitoring,
we also demonstrate an application to sequential anomaly detection.

The structure of the rest of the paper is as follows. Section \ref{sec:Forward-backward-inference}
reviews the conventional forward-backward algorithm for hidden Markov
model inference. Section \ref{sec:Density-ratio-inference} recasts
this procedure in terms of ratios of probability densities, and Section
\ref{sec:Density-ratio-estimation} describes how nonparametric estimates
of those density ratios can be calculated directly from training data
in both supervised and unsupervised settings, and how estimates of
several pairwise likelihood ratio functions can be obtained from a
concise set of parameters. Experimental results in Section \ref{sec:Experiments}
show the performance of the density ratio HMM methodology on synthetic
and real-world physiological monitoring data, showing striking improvements
compared to conventional parametric and nonparametric sequential inference
approaches.

A Matlab implementation and demo is available at \url{http://cit.mak.ac.ug/staff/jquinn/software/densityratioHMM.html}.

\section{Forward-backward inference\label{sec:Forward-backward-inference}}

Consider the estimation of a latent sequence $x_{1:T}=\left\{ x_{t}\in\mathcal{S}|t=1,\ldots,T\right\} $
from an observation sequence $\mathbf{y}_{1:T}=\left\{ \mathbf{y}_{t}\in\mathbb{R}^{d_{y}}|t=1,\ldots,T\right\} $.
The variable $x_{t}$ is assumed to have first order Markovian dynamics,
such that $p(x_{t}|x_{1:t-1})=p(x_{t}|x_{t-1})$, the values it can
take on are a discrete set of classes or `states' $\mathcal{S}=\left\{ 1,\ldots,S\right\} $,
and each observation is independently drawn from a fixed emission
distribution $p(\mathbf{y}_{t}|x_{t})$. Given a sequence $\mathbf{y}_{1:T}$
and knowledge of both $p(x_{t}|x_{t-1})$ and $p(\mathbf{y}_{t}|x_{t})$
and the initial state distribution $p(x_{1})$, the forward-backward
algorithm can be used to estimate the probability of each state at
each time frame, $p(x_{t}{=}i|\mathbf{y}_{1:T})$. We use the following
shorthand in describing the algorithm: $\mathbf{A}$ denotes a matrix
of state transition probabilities, such that $\mathbf{A}_{ij}=p(x_{t}{=}j|x_{t-1}{=}i)$
and $\pi$ is a vector of initial state probabilities such that $\pi_{i}=p(s_{1}{=}i)$.

The first stage of the algorithm involves recursively calculating
forward messages $\alpha_{t}(i)$ for each of the states $i\in\mathcal{S}$:
\begin{eqnarray}
\tilde{\alpha}_{i}(1) & = & \pi_{i}p(\mathbf{y}_{t}|x_{1}{=}i)\,,\label{eq:forwardinit}\\
\tilde{\alpha}_{i}(t) & = & \left[\sum_{j\in\mathcal{S}}\alpha_{j}(t-1)\mathbf{A}_{ji}\right]p(\mathbf{y}_{t}|x_{t}{=}i)\,,\label{eq:forward}\\
 &  & t=2,\ldots,T.\nonumber 
\end{eqnarray}
After each time step, it is conventional (but not necessary) to normalize
the forward messages, 
\begin{equation}
\alpha_{i}(t)=c_{t}\tilde{\alpha}_{i}(t)\,\mathrm{\,\,\, s.t.\,\,\,\,}\sum_{i\in\mathcal{S}}\alpha_{i}(t)=1\,.\label{eq:forwardnormalised}
\end{equation}
Without this normalization the algorithm is numerically unstable,
as the values of $\tilde{\alpha}_{i}$ otherwise become very small
over repeated iterations. The step is optional however because normalization
occurs anyway later in the procedure (in Eq. (\ref{eq:gamma})). The
forward messages can be interpreted as the posterior probability of
each state given observations up to that time frame, the process of
calculating this being known as \emph{filtering}. Note that the normalization
means that the absolute values of $p(\mathbf{y}_{t}|x_{t}{=}i)$ are
not directly significant for inference; we are ultimately interested
only in the relative magnitudes.

To carry out \emph{smoothing} (calculation of $p(x_{t}{=}i|\mathbf{y}_{1:T})$)
the backwards messages $\beta_{t}(i)$ must first be similarly calculated:
\begin{eqnarray}
\tilde{\beta}_{i}(T) & = & 1\,,\label{eq:backwardinit}\\
\tilde{\beta}_{i}(t) & = & \sum_{j\in\mathcal{S}}\mathbf{A}_{ij}p(\mathbf{y}_{t+1}|x_{t+1}{=}j)\beta_{j}(t+1)\,,\label{eq:backward}\\
 &  & t=T-1,\ldots,1,\nonumber 
\end{eqnarray}
also with an optional normalization step carried out after each iteration,
\begin{equation}
\beta{}_{i}(t)=c_{t}^{'}\tilde{\beta}_{i}(t)\,\mathrm{\,\,\, s.t.\,\,\,\,}\sum_{i\in\mathcal{S}}\beta_{i}(t)=1\,.\label{eq:backwardnormalised}
\end{equation}
The two types of messages are then combined to give the final result
\begin{equation}
\gamma_{i}(t)=p(x_{t}{=}i|\mathbf{y}_{1:T})=\frac{\alpha_{i}(t)\beta_{i}(t)}{\sum_{j\in\mathcal{S}}\alpha_{j}(t)\beta_{j}(t)}\,.\label{eq:gamma}
\end{equation}

The forward-backward algorithm therefore requires an explicit likelihood
model for every dynamical regime, is numerically unstable (requiring
message scaling or transforms in and out of log space to prevent underflow),
and loses information during normalization steps. We next describe
how a density ratio formulation overcomes these problems, by expressing
inference in terms of parameters which are more natural to the problem.

\section{Inference with density ratios\label{sec:Density-ratio-inference}}

In this section we rearrange the above inference equations in terms
of pairwise probability density ratios. Ratios of probabilities make
it particularly convenient to express Bayesian updates, as the ratio
of two posteriors is equal to simply the ratio of the priors multiplied
by the ratio of the likelihoods. In order to express the forward-backward
equations in this way, corresponding to the three types of values
$\alpha_{i},\beta_{i},\gamma_{i}$ we define three types of ratios
$\overrightarrow{r}_{i,j},\overleftarrow{r}_{i,j},r_{i,j}$ as
\begin{align*}
\overrightarrow{r}_{i,j}(t) & =\frac{\alpha_{i}(t)}{\alpha_{j}(t)}=\frac{p(x_{t}{=}i|\mathbf{y}_{1:t})}{p(x_{t}{=}j|\mathbf{y}_{1:t})}\,,\\
\overleftarrow{r}_{i,j}(t) & =\frac{\beta_{i}(t)}{\beta_{j}(t)}=\frac{p(x_{t}{=}i|\mathbf{y}_{t+1:T})}{p(x_{t}{=}j|\mathbf{y}_{t+1:T})}\,,\\
r_{i,j}(t) & =\frac{\gamma_{i}(t)}{\gamma_{j}(t)}=\frac{p(x_{t}{=}i|\mathbf{y}_{1:T})}{p(x_{t}{=}j|\mathbf{y}_{1:T})}\,.
\end{align*}
We also treat likelihoods in terms of density ratios, and define
\begin{align*}
w_{i,j}(\mathbf{y}_{t}) & =\frac{p(\mathbf{y}_{t}|x_{t}{=}i)}{p(\mathbf{y}_{t}|x_{t}{=}j)}\,.
\end{align*}
Using Eqs. (\ref{eq:forwardinit}-\ref{eq:forward}) we can derive
the following expressions for density ratio versions of the forward
equations:
\begin{align}
\overrightarrow{r}_{i,j}(1)= & \frac{\pi_{i}}{\pi_{j}}w_{i,j}(\mathbf{y}_{1})\,,\label{eq:ratioforwardinit}\\
\overrightarrow{r}_{i,j}(t)= & \frac{\left[\sum_{k\in\mathcal{S}}\alpha_{k}(t-1)\mathbf{A}_{ik}\right]p(\mathbf{y}_{t}|x_{t}{=}i)}{\left[\sum_{k'\in\mathcal{S}}\alpha_{k'}(t-1)\mathbf{A}_{jk'}\right]p(\mathbf{y}_{t}|x_{t}{=}j)}\nonumber \\
= & \sum_{k\in\mathcal{S}}\left[\sum_{k'\in\mathcal{S}}\frac{\mathbf{A}_{jk'}}{\mathbf{A}_{ik}}\overrightarrow{r}_{k',k}(t-1)\right]^{-1}w_{i,j}(\mathbf{y}_{t})\,\nonumber \\
 & t=2,\ldots,T\,.\label{eq:ratioforward}
\end{align}
Although these expressions are written in terms of all pairwise ratios,
implying that $S^{2}$ terms have to be calculated at each time frame,
in fact these ratios can be estimated with a concise set of parameters
as we demonstrate in Section \ref{sec:Density-ratio-estimation}. 

For backwards messages we use Eqs. (\ref{eq:backwardinit}-\ref{eq:backward})
in a similar way to obtain
\begin{align}
\overleftarrow{r}_{i,j}(T)= & 1\\
\overleftarrow{r}_{i,j}(t)= & \frac{\sum_{k\in\mathcal{S}}\mathbf{A}_{ki}p(\mathbf{y}_{t+1}|x_{t+1}{=}k)\beta_{k}(t+1)}{\sum_{k'\in\mathcal{S}}\mathbf{A}_{k'j}p(\mathbf{y}_{t+1}|x_{t+1}{=}k')\beta_{k'}(t+1)}\nonumber \\
= & \sum_{k\in\mathcal{S}}\left[\sum_{k'\in\mathcal{S}}\frac{\mathbf{A}_{k'j}}{\mathbf{A}_{ki}}\overleftarrow{r}_{k',k}(t+1)w_{k',k}(\mathbf{y}_{t+1})\right]^{-1}\,.\nonumber \\
 & t=T-1,\ldots,1\,.
\end{align}
The messages are finally combined simply with
\[
r_{i,j}(t)=\overrightarrow{r}_{i,j}(t)\overleftarrow{r}_{i,j}(t)\,.
\]
The filtering and smoothing probabilities, if needed, can be simply
calculated from the ratios with $\alpha_{i}(t)=\frac{\overrightarrow{r}_{i,S}(t)}{\sum_{j\in S}\overrightarrow{r}_{j,S}(t)}$
and $\gamma_{i}(t)=\frac{r_{i,S}(t)}{\sum_{j\in S}r_{j,S}(t)}$. This
completes the density ratio forward-backward algorithm. 

\begin{figure}
\begin{centering}
\includegraphics[width=0.4\textwidth]{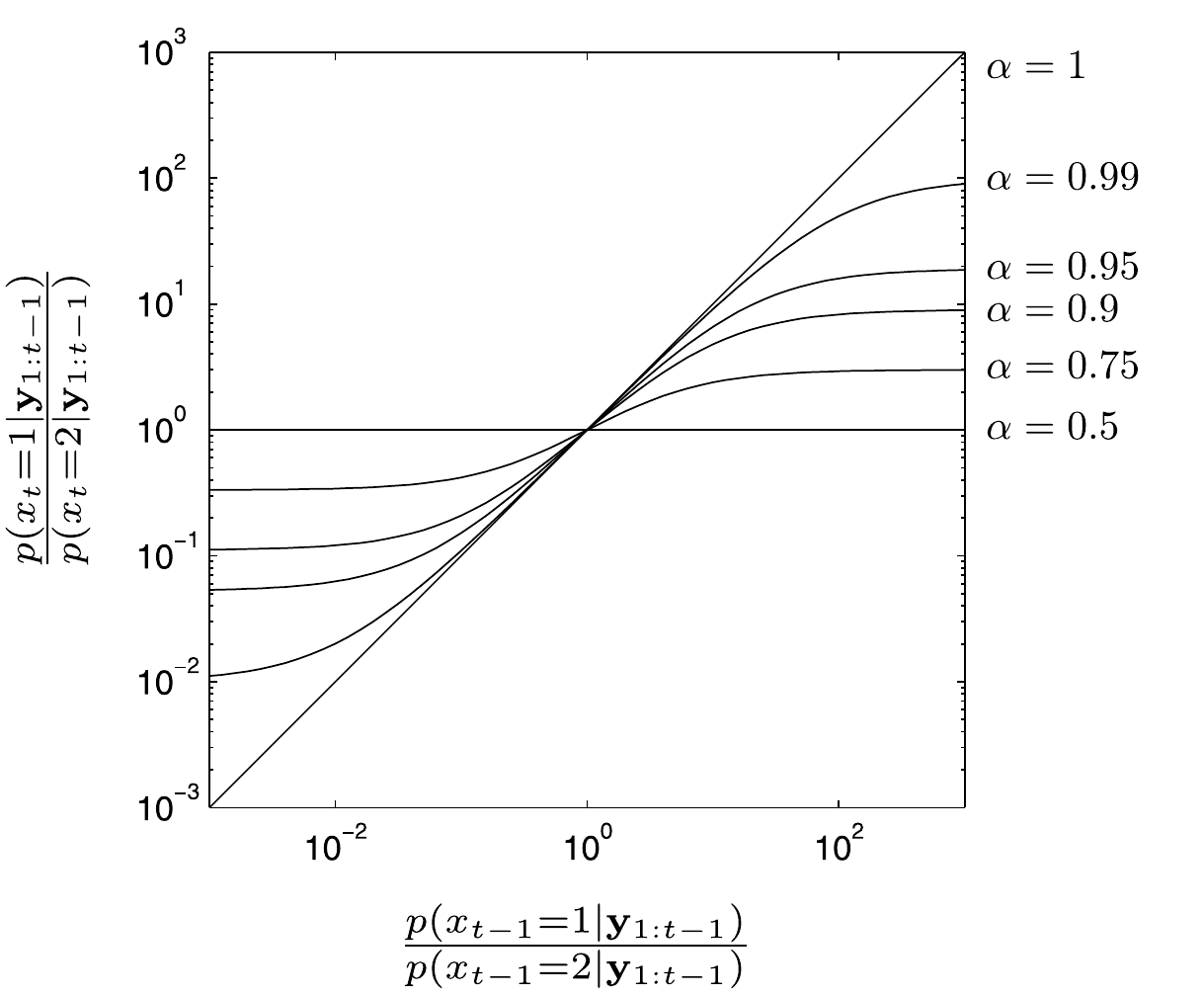}
\par\end{centering}

\caption{Forward probability ratio evolution function given different transition
matrices, for a 2-state HMM. See text for details.}
\label{fig:transitionratios}

\end{figure}

To provide some intuition about the ratio-based procedure we look
in some more detail at a two-state HMM, $\mathcal{S}=\left\{ 1,2\right\} $.
The forward equation in this simplified case reduces to:
\begin{align}
\overrightarrow{r}_{1,2}(t) & =\frac{\mathbf{A}_{11}\overrightarrow{r}_{1,2}(t-1)+\mathbf{A}_{12}}{\mathbf{A}_{21}\overrightarrow{r}_{1,2}(t-1)+\mathbf{A}_{22}}w_{1,2}(\mathbf{y}_{t})\,,\label{eq:twostateratioforward}
\end{align}
and the backward equation to
\begin{align}
\overleftarrow{r}_{1,2}(t) & =\frac{\mathbf{A}_{11}\overleftarrow{r}_{1,2}(t+1)w_{1,2}(\mathbf{y}_{t+1})+\mathbf{A}_{21}}{\mathbf{A}_{21}\overleftarrow{r}_{1,2}(t+1)w_{1,2}(\mathbf{y}_{t+1})+\mathbf{A}_{22}}\,.\label{eq:twostateratiobackward}
\end{align}
The first term of (\ref{eq:twostateratioforward}) specifies the evolution
from $\frac{p(x_{t-1}=1|\mathbf{y}_{1:t-1})}{p(x_{t-1}=2|\mathbf{y}_{1:t-1})}$
to $\frac{p(x_{t}=1|\mathbf{y}_{1:t-1})}{p(x_{t}=2|\mathbf{y}_{1:t-1})}$
in terms of the transition probabilities in \textbf{$\mathbf{A}$}.
Hence this is simply the ratio form of marginalizing out the transition
probabilities. Figure \ref{fig:transitionratios} shows this evolution
of the state probability ratio given transition matrices of the form
$\mathbf{A}=${\footnotesize $\left[\begin{array}{cc}
\alpha & 1-\alpha\\
1-\alpha & \alpha
\end{array}\right]$} for different values of $\alpha$. If $\alpha=0.5,$ then any past
information is lost; $\frac{p(x_{t}=1|\mathbf{y}_{1:t-1})}{p(x_{t}=2|\mathbf{y}_{1:t-1})}$
is always equal to 1, hence both states are equally likely \textit{a
priori} at time $t$. If $\alpha=1$, there can be no state transitions.
Inference is equivalent to accumulating evidence in favor of either
state 1 or 2 globally, and Eq. (\ref{eq:twostateratioforward}) reduces
to $\overrightarrow{r}_{1,2}(t)=\frac{\pi_{1}}{\pi_{2}}\prod_{t=1}^{t}w_{1,2}(\mathbf{y}_{t})$.
The ratio form of the Bayesian update given the observation $\mathbf{y}_{t}$
is simply multiplication by $w_{1,2}(\mathbf{y}_{t})$. 

The ratio formulation of the forward-backward algorithm is numerically
stable, and Figure \ref{fig:transitionratios} provides an intuition
of this; extreme ratio values tend to be mapped back to within a few
orders of magnitude of unity after considering transition probabilities.
Scaling of values to prevent underflow is therefore not necessary
with this method as it is in the conventional forward-backward algorithm.

Hence, given a way to estimate $w_{i,j}(\cdot)$ from training data,
it is possible to carry out inference in the HMM without ever needing
to calculate the individual likelihoods. These steps are mathematically
equivalent to the standard forward-backward algorithms, but do not
require any normalization step. If the observation distribution is
known exactly then the ratio formulation is equivalent and has no
advantage. However, when the observation distribution needs to be
approximated in some way -- which is almost always the case with complex
real-world data -- the parameterization in terms of $w_{i,j}(\cdot)$
is more natural to the sequential inference problem than attempting
to approximate each of the $p(\mathbf{y}_{t}|x_{t}{=}i)$ distributions
directly. We discuss ways in which the estimation of these ratios
and other parameters can be carried out in the next section.

\section{Parameter estimation\label{sec:Density-ratio-estimation}}

The parameters to be learned in the density ratio HMM model are the
likelihood ratio functions $w_{ij}(\mathbf{y})$, the transition matrix
$\mathbf{A}$ and initial class probabilities $\pi$. In this section
we first suggest effective methods for estimating $w_{ij}(\mathbf{y})$
which bypass the difficult step of estimating the densities $p(\mathbf{y}|x=i)$
and $p(\mathbf{y}|x=j)$ individually. In Section \ref{sub:Two-class-density-ratio}
we introduce an efficient method for carrying this out when there
are two classes to be modeled, an give a related method for several
classes in Section \ref{sub:Extension-to-multiple} which does not
need every pairwise ratio to be estimated individually. The estimation
of $\mathbf{A}$ and $\pi$ is then discussed in Section \ref{sec:other_params}.

\subsection{Two-class density ratio estimation\label{sub:Two-class-density-ratio}}

We begin by discussing how to estimate a likelihood ratio from data
in the case that there are only two states $i$ and $j$ in the model,
for example as would be needed to calculate (\ref{eq:twostateratioforward})
and (\ref{eq:twostateratiobackward}). There is just one ratio to
learn, since $w_{ij}=w_{ji}^{-1}$. Several techniques have been developed
for direct density ratio estimation \cite{Biometrika:Qin:1998,ICML:Bickel+etal:2007,Inbook:Gretton+etal:2009,IEEE-IT:Nguyen+etal:2010,AISM:Sugiyama+etal:2008,AISM:Sugiyama+etal:2012}.
We use a least squares approach here \cite{JMLR:Kanamori+etal:2009}
which yields a consistent estimator with very good computational efficiency.
The estimator is of the following form:
\[
\hat{w}_{ij}(\mathbf{y})=\boldsymbol{\theta}_{ij}^{\top}\bm{\phi}(\mathbf{y})\,,
\]
where
\[
\boldsymbol{\theta}_{ij}=(\theta_{ij,1},\ldots,\theta_{ij,B})^{\top}\in\mathbb{R}^{B}
\]
for some number of parameters $B$, and
\begin{equation}
\boldsymbol{\phi}(\mathbf{y})=(K(\mathbf{y},\mathbf{y}_{1}),\ldots,K(\mathbf{y,y}_{B}))^{\top}\in\mathbb{R}^{B}\label{eq:kernelbasis}
\end{equation}
is a vector of kernel basis functions. We can set $B=N$ to have a
kernel basis function at every training point, or for $B<N$ use some
random subset of the training points. In this work we use the squared
exponential kernel $K(\mathbf{y},\mathbf{y}')=\exp\left(\frac{-||\mathbf{y}-\mathbf{y}'||^{2}}{2\sigma^{2}}\right)$.

This model can be fitted using a squared loss objective function,
\[
J_{ij}(\boldsymbol{\theta}_{ij})=\frac{1}{2}\int\left(\hat{w}_{ij}(\mathbf{y})-\frac{p(\mathbf{y}|x{=}i)}{p(\mathbf{y}|x{=}j)}\right)^{2}p(\mathbf{y}|x{=}j)d\mathbf{y}\,.
\]
Expanding the squared term we obtain
\begin{align*}
J_{ij}(\boldsymbol{\theta}_{ij})= & \frac{1}{2}\int\hat{w}_{ij}(\mathbf{y}){}^{2}p(\mathbf{y}|x{=}j)d\mathbf{y}\\
 & -\int\hat{w}_{ij}(\mathbf{y})p(\mathbf{y}|x{=}i)d\mathbf{y}+C\,,
\end{align*}
where $C$ is a constant term that does not depend on any of the $\boldsymbol{\theta}_{ij}$
values. Empirically, we can approximate the expectations by sample
averages. Ignoring the constant $C$, factor $1/N$ and including
an $\ell_{2}$-regularizer, we have the following training criterion:
\[
\widehat{J}_{ij}(\boldsymbol{\theta}_{ij})=\frac{1}{2}\boldsymbol{\theta}_{ij}^{\top}\mathbf{\boldsymbol{\Phi}}^{\top}\mathbf{M}_{j}\boldsymbol{\Phi}\boldsymbol{\theta}_{ij}-\boldsymbol{\theta}_{ij}^{\top}\boldsymbol{\Phi}\mathbf{m}_{i}+\frac{\rho}{2}||\theta_{i}||^{2},
\]
where $\boldsymbol{\Phi}=\left(\boldsymbol{\phi}(\mathbf{y}_{1}),\ldots,\boldsymbol{\phi}(\mathbf{y}_{N})\right)^{\top}$,
$\mathbf{m}_{i}$ is a column vector indicating membership of class
$i$ such that the $j$th element is one if $x_{j}=i$ and zero otherwise,
and $\mathbf{M}_{j}$ is a square matrix with $\mathbf{m}_{j}$ along
the diagonal and other entries set to zero. $\widehat{J}_{i}(\boldsymbol{\theta}_{i})$
is minimized by
\begin{equation}
\widehat{\boldsymbol{\theta}}_{ij}=\left(\boldsymbol{\Phi}^{\top}\mathbf{M}_{j}\boldsymbol{\Phi}+\rho\mathbf{I}_{B}\right)^{-1}\boldsymbol{\Phi}^{\top}\boldsymbol{m}_{i}\,.\label{eq:theta_hat-1}
\end{equation}
We select $\rho$ and $\sigma$ with cross validation. Because of
the nature of the estimator, it is sometimes possible to obtain negative
values for $\hat{w}_{ij}(\mathbf{y})$. We simply round up negative
estimates to zero in such cases, which does not affect the consistency
of the estimator \cite{kanamori12statistical}. This least-squares
approach is very fast to compute in practice, finding a global optimum
in a single step with no iterative parameter search required.

\subsection{Extension to multiple classes\label{sub:Extension-to-multiple}}

In problems where $S>2$, it would be possible to simply learn several
pairwise likelihood ratio functions. Because $w_{ij}=w_{ji}^{-1}$
and $w_{ii}=1$, this would entail learning $\frac{S^{2}-S}{2}$ ratio
functions. Fewer still need to be estimated if relationships of the
form $w_{ij}=w_{ik}/w_{jk}$ are used, though this `ratio of ratios'
may become unstable when the denominator $w_{jk}$ is close to zero.

For multiple-class ratio estimation another approach is to directly
estimate conditional probabilities of class $x_{t}$ given i.i.d.
samples $\left\{ \mathbf{y}^{(1)},\ldots,\mathbf{y}^{(N)}\right\} $.
Using the fact that $p(x{=}i|\mathbf{y})=\frac{p(\mathbf{y}|x{=}i)p(x{=}i)}{\sum_{j\in\mathcal{S}}p(\mathbf{y}|x{=}j)p(x{=}j)}$,
likelihood ratios in our problem can be estimated as 
\begin{equation}
\hat{w}_{i,j}(\mathbf{y})=\frac{p(\mathbf{y}|x{=}i)}{p(\mathbf{y}|x{=}j)}=\frac{p(x{=}j)}{p(x{=}i)}\frac{p(x{=}i|\mathbf{y})}{p(x{=}j|\mathbf{y})}\,.\label{eq:posteriorratio}
\end{equation}
This is similar in principle to the scheme of using classifiers to
calculate likelihoods in standard HMMs discussed in \citet{punyakanok01classifiers}. 

A standard approach to calculating the $p(x_{t}{=}i|\mathbf{y})$
terms is kernel logistic regression. This can be computationally demanding
for large datasets though, so we use a least squares procedure \cite{sugiyama12LSPC}
similar in form to the two-class estimator we introduced in the previous
section, and which is known to give comparable accuracy to kernel
logistic regression but requiring far less training time. 

We construct functions $q(x{=}i|\mathbf{y},\mathbf{\boldsymbol{\theta}}_{i})$
to estimate $p(x_{t}{=}i|\mathbf{y})$, defined as
\[
q(x{=}i|\mathbf{y},\mathbf{\boldsymbol{\theta}}_{i})=\boldsymbol{\theta}_{i}^{\top}\bm{\phi}(\mathbf{y})
\]
where
\[
\boldsymbol{\theta}_{i}=(\theta_{i,1},\ldots,\theta_{i,B})^{\top}\in\mathbb{R}^{B}
\]
and $\boldsymbol{\phi}(\mathbf{y})$ is the same as in Eq. (\ref{eq:kernelbasis}).
The squared loss term in this case is
\[
J_{i}(\boldsymbol{\theta}_{i})=\frac{1}{2}\int\left(q(x{=}i|\mathbf{y},\mathbf{\boldsymbol{\theta}}_{i})-p(x{=}i|\mathbf{y})\right)^{2}p(\mathbf{y})d\mathbf{y}\,.
\]
Expanding and using $p(x|\mathbf{y})=p(\mathbf{y}|x)p(x)/p(\mathbf{y})$
we obtain
\begin{align*}
J_{i}(\boldsymbol{\theta}_{i})= & \frac{1}{2}\int q(x{=}i|\mathbf{y},\mathbf{\boldsymbol{\theta}}_{i})^{2}p(\mathbf{y})d\mathbf{y}\\
 & -\int q(x{=}i|\mathbf{y},\mathbf{\boldsymbol{\theta}}_{i})p(\mathbf{y}|x{=}i)p(x{=}i)d\mathbf{y}+C\,,
\end{align*}
which can be empirically approximated to give the following training
criterion:
\[
\widehat{J}_{i}(\boldsymbol{\theta}_{i})=\frac{1}{2}\boldsymbol{\theta}_{i}^{\top}\mathbf{\boldsymbol{\Phi}}^{\top}\boldsymbol{\Phi}\boldsymbol{\theta}_{i}-\boldsymbol{\theta}_{i}^{\top}\boldsymbol{\Phi}\mathbf{m}_{i}+\frac{\rho}{2}||\theta_{i}||^{2}\,.
\]
$\widehat{J}_{i}(\boldsymbol{\theta}_{i})$ is minimized by
\begin{equation}
\widehat{\boldsymbol{\theta}}_{i}=\left(\boldsymbol{\Phi}^{\top}\boldsymbol{\Phi}+\rho\mathbf{I}_{B}\right)^{-1}\boldsymbol{\Phi}^{\top}\boldsymbol{m}_{i}\,,\label{eq:theta_hat}
\end{equation}
which is essentially kernel ridge regression. As in the two-class
case, we select $\rho$ and $\sigma$ with cross validation. Also
as before we round up negative estimates to zero, $q(x{=}i|\mathbf{y},\widehat{\mathbf{\boldsymbol{\theta}}}_{i})=\max\left(0,\widehat{\boldsymbol{\theta}}_{i}^{\top}\bm{\phi}(\mathbf{y})\right)$,
though we note that that negative estimates are unlikely when the
number of training samples is large enough \cite{sugiyama12LSPC}.

Now using Eq. (\ref{eq:posteriorratio}) we are able to calculate
likelihood ratios for any pair of states using
\[
\widehat{w}_{i,j}(t)=\frac{n_{i}}{n_{j}}\,\frac{q(x_{t}{=}i|\mathbf{y}_{t},\widehat{\boldsymbol{\theta}}_{i})}{q(x_{t}{=}j|\mathbf{y}_{t},\widehat{\boldsymbol{\theta}}_{j})}\,,
\]
where $n_{i}$ is the number of training samples labeled $x_{t}=i$.
Just as in the two-class case, Eq. (\ref{eq:theta_hat}) can be computed
very quickly.

\subsection{Setting other parameters\label{sec:other_params}}

We now discuss setting parameters $\mathbf{A},\pi$ in the density
ratio HMM given training data, first when states $x_{1:T}$ are available
and second as an unsupervised learning problem when we only have observation
sequences $\mathbf{y}_{1:T}$.

If $x_{1:T}$ is observed in training data, $p(x_{t}{=}i|x_{t-1}{=}j)$
and $p(x_{1}{=}i)$ can be estimated directly by frequency. The process
of setting the remaining parameters $\mathbf{A},\pi$ in the case
that the labels $x_{1:T}$ are present in the training data is therefore
identical to the standard HMM; we refer the reader to the details
in \citet{rabiner1989tutorial}.

For the unsupervised case in which only the observation sequence $\mathbf{y}_{1:T}$
is available for training, learning can be carried out by iterating
between (1) a likelihood-maximization step to update $\mathbf{A},\pi,\left\{ \boldsymbol{\theta}_{i}\right\} $
given estimates of $p(x_{1:T}|\mathbf{y}_{1:T})$ and (2) an expectation
step, the inference procedure given in §\ref{sec:Density-ratio-inference}.
To do this some initial estimate $\widehat{x}_{1:T}$ is required
as a starting point, which could be obtained by running a standard
non-dynamic clustering procedure such as $k$-means on the observations.
The iterations are continued until convergence or some limit on the
number of cycles is reached. For unsupervised learning we require
one alteration to the parameter estimation procedure for $w_{ij}(\cdot)$
in order to accommodate soft estimates of state probabilities $\gamma_{i}$
rather than hard labels. The weighted version of Eq. (\ref{eq:theta_hat})
is 
\[
\widehat{\boldsymbol{\theta}}_{i}=\left(\boldsymbol{\Phi}^{\top}\mathbf{\boldsymbol{\Gamma}}^{(i)}\boldsymbol{\Phi}+\rho\mathbf{I}_{B}\right)^{-1}\boldsymbol{\Phi}^{\top}\left[\gamma_{i}(1),\ldots,\gamma_{i}(B)\right]^{\top}
\]
where $\boldsymbol{\Gamma}^{(i)}$ is a $B\times B$ diagonal matrix
such that $\boldsymbol{\Gamma}_{j,j}^{(i)}=\gamma_{i}(j)$. The maximization
step updates for $\mathbf{A},\pi$ given $\gamma_{i}$ are again identical
to the standard HMM case \cite{rabiner1989tutorial}.

\section{Experiments\label{sec:Experiments}}

We now give experimental results using the above methods when applied
to both synthetic and real-world datasets.\begin{figure*}[t] 
\begin{centering} 
\includegraphics[bb=120bp 0bp 760bp 283bp,width=0.9\textwidth]{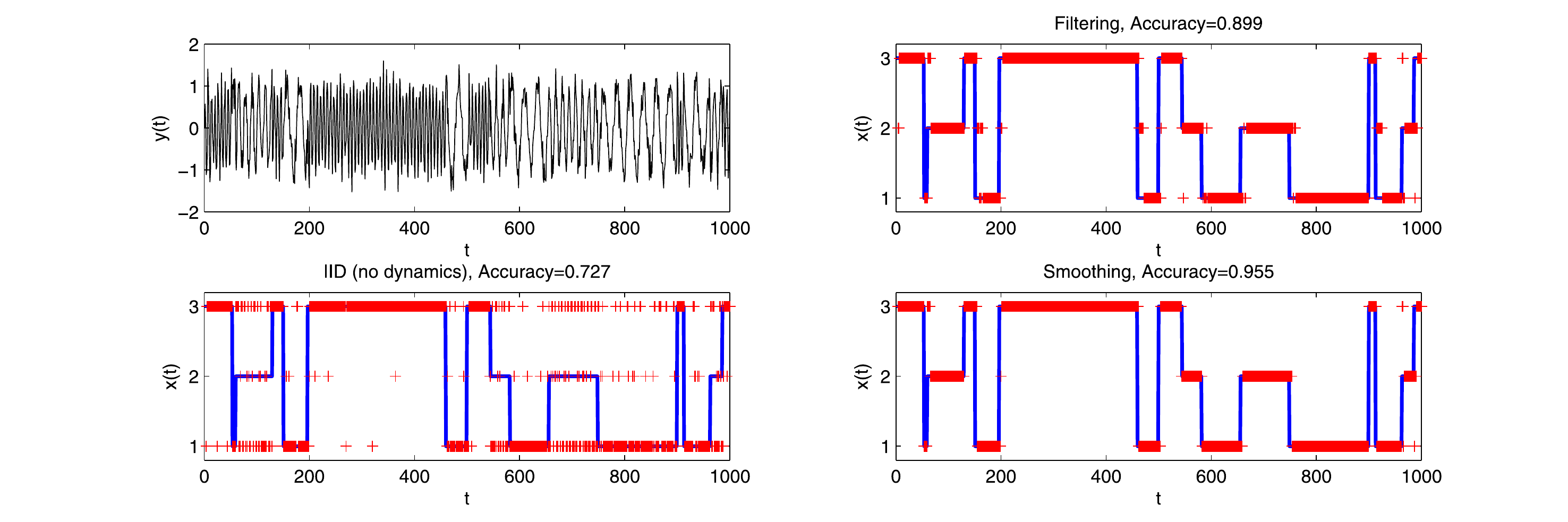} \par
\end{centering}
\caption{Inference example on noisy sine wave sequence. The top left panel shows a sampled sequence, which switches randomly between three dynamic regimes (sine waves of three different frequencies with Gaussian noise added). In the bottom left panel, the blue line shows the true value of $x_{t}$, and the red crosses show the estimated most likely value given density ratio estimates independently at each time frame. The top right panel shows the inference improving when forward dynamics are incorporated; the bottom right panel shows further improvement when both forward and backward dynamics are incorporated.}
\label{fig:inferencetoyexample} 
\end{figure*}

\subsection{Sequential classification on toy data}

\begin{figure*}[t]
\begin{centering} 
\includegraphics[bb=50bp 0bp 800bp 227bp,width=1\textwidth]{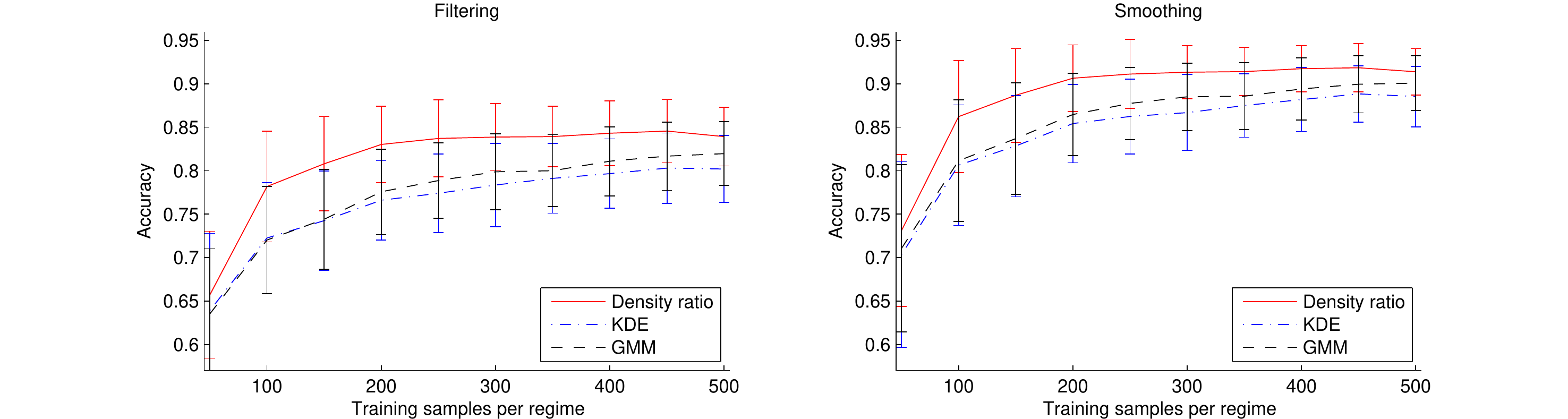} \par
\end{centering}
\caption{Accuracy of state estimates in noisy switching sine wave problem, mean and standard deviation over 500 runs per training sample size.} 
\label{fig:toyinferenceresults}
\end{figure*}

To give an illustration of HMM inference using density ratio estimation,
we generated data from a simple switching model with three dynamical
regimes. The transition probabilities in the model were set to $\mathbf{A}=${\scriptsize $\left[\begin{array}{ccc}
.98 & .01 & .01\\
.01 & .98 & .01\\
.01 & .01 & .98
\end{array}\right]$} with initial state probabilities $\pi=\left[\frac{1}{3},\frac{1}{3},\frac{1}{3}\right]$,
and these parameters were then used to generate sequences $x_{1:T}$.
Scalar sequences $y_{1:T}$ were then generated conditioned on $x_{t}$
as follows:
\[
y_{t}=\begin{cases}
\sin(0.2t)+\eta_{t} & x_{t}=1\\
\sin(0.4t)+\eta_{t} & x_{t}=2\\
\sin(0.6t)+\eta_{t} & x_{t}=3
\end{cases}
\]
with $\eta_{t}\sim\mathcal{N}(0;0.25)$. A sample sequence is shown
in Figure \ref{fig:inferencetoyexample} (top left). To construct
the vector sequence $\mathbf{y}_{1:T}$ used for testing inference,
we took subsequences of $y_{1:T}$ using a sliding window of length
$d_{y}$ (using $d_{y}=4$ for this example) such that $\mathbf{y}_{t}=y_{t-d_{y}+1:t}$,
for $t\geq d_{y}$.

Figure \ref{fig:inferencetoyexample} (bottom left) shows the results
of using density ratios at each time step to find the most likely
class independently at each time step, by finding $\argmax_{i}\left(\sum_{j\in\mathcal{S}}r_{i,j}(\mathbf{y}_{t})\right)$.
This is equivalent to filtering inference using a uniform transition
matrix $\mathbf{A}_{ij}=\frac{1}{S}$, and gives an idea of how informative
the subsequences $\mathbf{y}_{t}$ are about $x_{t}$ at individual
time frames when no dynamical information is incorporated. In this
plot the blue line shows the true values of $x_{1:T}$ in the sample
sequence, and the red crosses show the MAP estimates. The top right
panel shows MAP estimates with filtering inference using the forward
equations only. The bottom right panel shows MAP estimates with smoothing
inference using both forward and backward steps.

Inference results were compared to those from two alternative models.
The first was an alternative nonparametric hidden Markov model, using
kernel density estimation (using a squared exponential kernel, with
parameters chosen by cross-validation) to model $p(\mathbf{y}_{t}|x_{t}{=}i)$
for each $i\in\mathcal{S}$, as proposed in \citet{piccardi07kdeHMM}.
The second was the conventional Gaussian mixture model approach to
modeling the observation density of each regime \cite{rabiner1989tutorial},
with the number of components of the mixture selected in each case
being that which minimized the Bayesian Information Criterion (BIC)
on training data. Training sequences were randomly generated from
the above switching noisy sine wave example, of lengths between 50
and 500 subsequences for each of the three dynamic regimes. Test sequences
of length 1000 were also generated as above. The density ratio HMM,
KDE-HMM and GMM-HMM were trained and applied to the test sequence.
The accuracy, calculated as the proportion of time frames for which
the MAP estimate $\max_{i}p(x_{t}{=}i|\mathbf{y}_{1:T})$ was equal
to the true simulated state $x_{t}$, was calculated for both methods.
This was repeated 500 times for each training sample size. Figure
\ref{fig:toyinferenceresults} shows the mean and standard deviation
of accuracy for filtering and smoothing with all three methods. The
density ratio HMM has consistently higher average accuracy than the
other methods, particularly for small training set sizes.

\subsection{Physiological monitoring}

We next evaluated the density ratio HMM using time series of physiological
measurements from premature infants receiving intensive care. The
dataset we used consists of 24 hour sequences of monitoring data from
15 babies, a total of 360 hours of data with measurements taken at
one second intervals. The measurements are of vital signs (heart rate,
blood pressures, temperatures, blood gas concentrations) and environmental
measurements (incubator temperature and humidity). The data is annotated
with the occurrences of four common phenomena: bradycardia (temporary
slowing or stopping of the heart), opening of the incubator, the taking
of a blood sample, and disconnection of the core temperature probe.
Any period in the data during which there was some clinically significant
change not covered by one of the four conditions above was annotated
as a fifth class. Finally, for each baby a period of around 10 minutes
was annotated as `normal', i.e. representative of that baby's baseline
physiology. The data is publicly available and described in \citet{quinn09physiological}.

We first trained a set of two-class density ratio HMMs with this data,
treating each state in the annotation as a separate inference problem.
To train the bradycardia model for a particular baby, for example,
we would take the period for that baby annotated as normal as training
data for the first state, and periods annotated as bradycardia from
other babies as the training data for the second state.

We compared the output of the density ratio HMM to that of a factorial
hidden Markov model (FHMM) and factorial switching linear dynamical
system (SLDS), recreating the evaluation of \citet{quinn09physiological}.
As the problem associated with this dataset is real-time patient monitoring,
we applied filtering inference only. The latter methods were developed
using extensive domain knowledge as to the physical processes underlying
the observations. The evaluation was done using 3-fold cross validation
on the set of 15 sequences. Area under ROC curve (AUC) and equal error
rate (EER) for each of the classes in the annotations were calculated
for the three methods, shown in Table \ref{tab:ICUresults}. The density
ratio HMM gives either equivalent or superior results in all cases,
for example achieving a 17\% increase in AUC and 13\% decrease in
EER for detection of temperature probe disconnection compared to the
next best method.  

\begin{table}[t]
\begin{centering}
\begin{tabular}{ccccc}
\toprule 
 &  & {\small FHMM} & {\small SLDS} & {\small DR-HMM}\tabularnewline
\midrule
\multirow{2}{*}{{\small Bradycardia}} & {\scriptsize AUC} & .66 & .88 & \textbf{.92}\tabularnewline
 & {\scriptsize EER} & .37 & .25 & \textbf{.13}\tabularnewline
\midrule 
\multirow{2}{*}{{\small Incu. Open}} & {\scriptsize AUC} & .78 & .87 & \textbf{.88}\tabularnewline
 & {\scriptsize EER} & .25 & .17 & .17\tabularnewline
\midrule 
\multirow{2}{*}{{\small Blood Sample}} & {\scriptsize AUC} & .82 & .96 & .96\tabularnewline
 & {\scriptsize EER} & .20 & .14 & \textbf{.05}\tabularnewline
\midrule 
\multirow{2}{*}{{\small Temp. Probe}} & {\scriptsize AUC} & .74 & .77 & \textbf{.94}\tabularnewline
 & {\scriptsize EER} & .32 & .23 & \textbf{.10}\tabularnewline
\midrule
\multirow{2}{*}{Abnormal} & {\scriptsize AUC} & - & .69 & \textbf{.75}\tabularnewline
 & {\scriptsize EER} & - & .36 & \textbf{.31}\tabularnewline
\bottomrule
\end{tabular}\caption{Classification accuracy on neonatal intensive care unit time series
data, for occurrences of bradycardia, opening of the incubator, blood
sampling, temperature probe disconnection and ``other significant
deviation from normal dynamics''. FHMM denotes the factorial hidden
Markov model, SLDS denotes a switching linear dynamical system, and
DR-HMM denotes the density ratio hidden Markov model. The highest
AUC and lowest EER are highlighted in bold.}

\par\end{centering}

\label{tab:ICUresults}
\end{table}

\subsubsection*{Anomaly detection}

We also constructed an anomaly detection model for each test sequence,
to assess the ability of this framework to identify any clinically
significant deviation from known types of physiological variation.
Using the multiple-class density ratio estimation procedure in §\ref{sub:Extension-to-multiple},
this can be modeled quite easily%
\footnote{An outlier detection method which could be used with the two-class
ratio estimation method in §\ref{sub:Two-class-density-ratio} is
described in \citet{JMLR:Kanamori+etal:2009}. %
}. 

Assuming that observations from outlier classes might be present in
test data, we use $x_{t}{=}\ast$, $\ast\notin\mathcal{S}$ to denote
any such class at time $t$. The method we propose for discriminating
outliers from inliers is similar in essence to the one-class support
vector machine \cite{scholkopf1999sv} and the kernel Fisher discriminant
method for outlier detection \cite{roth2006kernel}. These methods
are based on the assumption that outliers occupy low-density regions
of the data space and that a kernel model can be used to characterize
the high-density regions given training data. Any given significance
threshold can then be used to separate the inlier and outlier level
sets. 

We estimate the conditional probability of an outlier $p(x{=}\ast|\mathbf{y},\mathbf{\boldsymbol{\theta}}_{i})$
with
\begin{equation}
q(x{=}*|\mathbf{y},\mathbf{\boldsymbol{\theta}}_{*})=1-\boldsymbol{\theta}_{*}^{\top}\bm{\phi}(\mathbf{y})\,.\label{eq:abnormalmodel}
\end{equation}
The problem of identifying outliers can then be equated with learning
$\boldsymbol{\theta}_{*}$ such that Eq. (\ref{eq:abnormalmodel})
is close to zero when $\mathbf{y}$ is within a region in which training
data has high density, and is close to one anywhere else. To achieve
this we minimize the following loss function:
\begin{align}
J_{2}(\boldsymbol{\theta}_{2}) & =\frac{1}{2}\int\left(1-\boldsymbol{\theta}_{2}^{\top}\bm{\phi}(\mathbf{y})\right)^{2}p(\mathbf{y})d\mathbf{y}+\frac{\rho}{2}||\boldsymbol{\theta}_{2}||^{2}\,.\label{eq:anomaly_objective}
\end{align}
The solution to this is given by
\begin{align*}
\widehat{\boldsymbol{\theta}}_{*} & =\left(\boldsymbol{\Phi}^{\top}\boldsymbol{\Phi}+\rho\mathbf{I}_{B}\right)^{-1}\sum_{t=1}^{N}\boldsymbol{\phi}(\mathbf{y}_{t})\\
 & =\left(\boldsymbol{\Phi}^{\top}\boldsymbol{\Phi}+\rho\mathbf{I}_{B}\right)^{-1}\sum_{i\in\mathcal{S}}\boldsymbol{\Phi}^{\top}\boldsymbol{m}_{i}\\
 & =\sum_{i\in\mathcal{S}}\widehat{\boldsymbol{\theta}}_{i}\,.
\end{align*}
Therefore
\[
q(x{=}*|\mathbf{y},\widehat{\mathbf{\boldsymbol{\theta}}}_{1},\ldots,\widehat{\boldsymbol{\theta}}_{S})=1-\sum_{i\in\mathcal{S}}\widehat{\boldsymbol{\theta}}_{i}^{\top}\bm{\phi}(\mathbf{y})\,.
\]
The parameters learned to model the inlier classes can therefore also
be used for outlier detection. 

To apply this to the problem of anomaly detection in the physiological
monitoring dataset, we trained inlier class parameters using the reference
period of the test sequence annotated as normal as well as any subsequence
of training data annotated as bradycardia, incubator opening, blood
sampling or temperature probe disconnection. Inference of $p(x_{1:T}=*|\mathbf{y}_{1:T})$
in test data therefore gave an estimate of whether any significant
physiological changes not consistent with any of the known dynamical
regimes were occurring.

We compared this to the anomaly detection method based on the SLDS
in \citet{quinn09physiological}. Again using 3-fold cross-validation
on the 360 hours of annotated data, the performance of our method
was found to be superior to that of the SLDS, with 6\% increase in
AUC and 5\% decrease in EER.

\section{Conclusions\label{sec:Conclusions}}

In this paper we have demonstrated that direct density ratio estimation
methods can be applied to sequential inference problems, improving
the discriminative capability of HMMs without losing the probabilistic
interpretation of such models. We believe this method is particularly
effective when no parametric class-conditional distribution of observations
is known a priori, which is usually the case in real-world problems.
As density ratio estimation is a growing field, having already been
successfully applied to various problems in statistical inference
such as covariate shift and outlier detection, the ideas in this paper
make further advances in the field directly applicable to sequence
modeling. It would be possible to apply this same principle to several
other related sequential probabilistic models, another significant
direction for future work. 

\bibliographystyle{plain}
\bibliography{densityratioHMM_preprint.bbl}

\end{document}